\documentclass{article} 
\usepackage{nips12submit_e,times}
\usepackage{caption}
\usepackage{amsfonts}
\usepackage{amsmath}
\usepackage{amssymb}
\usepackage{CJK}
\usepackage{graphicx}
\usepackage{algorithm}
\usepackage{algorithmic}
\usepackage{graphics}
\usepackage{multirow}
\usepackage[lofdepth,lotdepth]{subfig}

\def\shortcite{\cite}
\def\nipsfinalcopy{\nipsfinaltrue}

\title{Joint Space Neural Probabilistic Language Model for Statistical Machine Translation}

\author{
Tsuyoshi Okita
\\
Dublin City University\\
Glasnevin, Dublin 9\\
Ireland\\
\texttt{tokita@computing.dcu.ie} \\
}

\nipsfinalcopy 

\begin{document}

\maketitle

\begin{abstract}
A neural probabilistic language model (\textsc{NPLM}) provides an idea to
achieve the better perplexity than n-gram language model and their smoothed
language models.  This paper investigates application area in bilingual NLP,
specifically Statistical Machine Translation (SMT). We focus on the
perspectives that \textsc{NPLM} has potential to open the possibility to
complement potentially `huge' monolingual resources into the
`resource-constraint' bilingual resources. We introduce an ngram-HMM language
model as \textsc{NPLM} using the non-parametric Bayesian construction.  In
order to facilitate the application to various tasks, we propose the joint
space model of ngram-HMM language model. We show an experiment of system
combination in the area of SMT. One discovery was that our treatment of noise
improved the results 0.20 BLEU points if \textsc{NPLM} is trained in
relatively small corpus, in our case 500,000 sentence pairs, which is often
the case due to the long training time of \textsc{NPLM}.
\end{abstract}

\section{Introduction}

A neural probabilistic language model (\textsc{NPLM})
\cite{Bengio00,Bengio05} and the distributed representations
\cite{Hinton86} provide an idea to achieve the better perplexity than
n-gram language model \cite{Stolcke02} and their smoothed language
models \cite{Kneser95,Chen98,Teh06}. Recently, the latter one,
i.e. smoothed language model, has had a lot of developments in the
line of nonparametric Bayesian methods such as hierarchical Pitman-Yor
language model (HPYLM) \cite{Teh06} and Sequence Memoizer (SM)
\cite{Wood09,Gasthaus10}, including an application to SMT
\cite{Okita10e,Okita11,Okita11c}. A \textsc{NPLM} considers the
representation of data in order to make the probability distribution
of word sequences more compact where we focus on the similar
semantical and syntactical roles of words. For example, when we have
two sentences {\sl ``The cat is walking in the bedroom''} and {\sl ``A
  dog was running in a room''}, these sentences can be more compactly
stored than the n-gram language model if we focus on the similarity
between (the, a), (bedroom, room), (is, was), and (running,
walking). Thus, a \textsc{NPLM} provides the semantical and
syntactical roles of words as a language model. A \textsc{NPLM} of
\shortcite{Bengio00} implemented this using the multi-layer neural
network and yielded 20\% to 35\% better perplexity than the language
model with the modified Kneser-Ney methods \cite{Chen98}.

There are several successful applications of \textsc{NPLM}
\cite{Schwenk07,Collobert08,Schwenk10,Collobert11a,Collobert11b,Deschacht12,Schwenk12}.
First, one category of applications include POS tagging, NER tagging,
and parsing \cite{Collobert11b,Bordes11}. This category uses the
features provided by a \textsc{NPLM} in the limited window
size.
It is often the case that there is no such long range effects that the
decision cannot be made beyond the limited windows which requires to
look carefully the elements in a long distance. Second, the other
category of applications include Semantic Role Labeling (SRL) task
\cite{Collobert11b,Deschacht12}. This category uses the features
within a sentence. A typical element is the predicate in a SRL task
which requires the information which sometimes in a long distance but
within a sentence. Both of these approaches do not require to obtain
the best tag sequence, but these tags are independent.  Third, the
final category includes MERT process \cite{Schwenk10} and possibly
many others where most of them remain undeveloped.  The objective of
this learning in this category is not to search the best tag for a
word but the best sequence for a sentence. Hence, we need to apply the
sequential learning approach.
Although most
of the applications described in
\cite{Collobert08,Collobert11a,Collobert11b,Deschacht12} are
monolingual tasks, the application of this approach to a bilingual
task introduces really astonishing aspects, which we can call
``creative words'' \cite{Veale12}, automatically into the traditional
resource constrained SMT components. For example, the training corpus
of word aligner is often strictly restricted to the given parallel
corpus. However, a \textsc{NPLM} allows this training with huge
monolingual corpus. Although most of this line has not been even
tested mostly due to the problem of computational complexity of
training \textsc{NPLM}, \shortcite{Schwenk12} applied this to MERT
process which reranks the n-best lists using \textsc{NPLM}. This paper
aims at different task, a task of system combination
\cite{Bangalore02,Matusov06,Tromble08,Du09b,DeNero09,Okita2012}. This
category of tasks employs the sequential method such as Maximum A
Posteriori (MAP) inference (Viterbi decoding)
\cite{Koller09,Sontag10,Murphy12} on Conditional Random Fields (CRFs)
/ Markov Random Fields (MRFs). 

Although this paper discusses an ngram-HMM language model which we
introduce as one model of \textsc{NPLM} where we borrow many of the
mechanism from infinite HMM \cite{VanGael09} and hierarchical
Pitman-Yor LM \cite{Teh06}, one main contribution would be to show one
new application area of \textsc{NPLM} in SMT. Although several
applications of \textsc{NPLM} have been presented, there have been no
application to the task of system combination as far as we know.

The remainder of this paper is organized as follows.  Section 2
describes ngram-HMM language model while Section 3 introduces a joint
space model of ngram-HMM language model.  In Section 4, our intrinsic
experimental results are presented, while in Section 5 our extrinsic
experimental results are presented. We conclude in Section 5.

\section{Ngram-HMM Language Model}

\begin{figure}[h]
\begin{center}
\includegraphics[width=12cm]{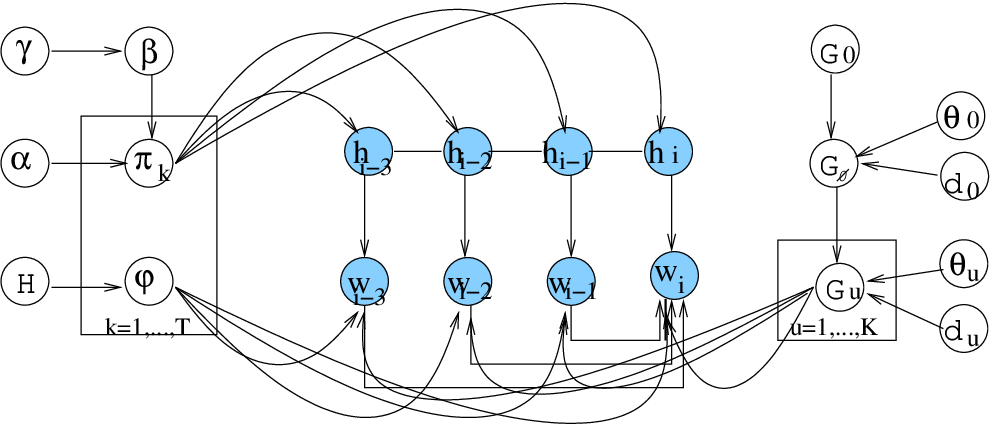}
\end{center}
\caption{Figure shows a graphical representation of the 4-gram HMM language model.}
\label{ngramHMMFigure}
\end{figure}

\paragraph{Generative model} 
Figure \ref{ngramHMMFigure} depicted an example of ngram-HMM language model,
i.e. 4-gram-HMM language model in this case, in blue (in the center).  We
consider a Hidden Markov Model (HMM) \cite{Rabiner89,Gharamani01,Beal03} of
size $K$ which emits n-gram word sequence $w_i,\ldots,w_{i-K+1}$ where
$h_i,\ldots,h_{i-K+1}$ denote corresponding hidden states.  The arcs from
$w_{i-3}$ to $w_{i}$, $\cdots$, $w_{i-1}$ to $w_{i}$ show the back-off
relations appeared in language model smoothing, such as Kneser-Ney smoothing
\cite{Kneser95}, Good-Turing smoothing \cite{Good53}, and hierarchical
Pitman-Yor LM smoothing \cite{Teh06}.

In the left side in Figure \ref{ngramHMMFigure}, we place one
Dirichlet Process prior $\mbox{DP}(\alpha,\mbox{H})$, with
concentration parameter $\alpha$ and base measure $H$, for the
transition probabilities going out from each hidden state. This
construction is borrowed from the infinite HMM
\cite{Beal03,VanGael09}. The observation likelihood for the hidden
word $h_t$ are parameterized as in $w_t|h_t \sim F(\phi_{h_t})$ since
the hidden variables of HMM is limited in its representation power
where $\phi_{h_t}$ denotes output parameters.  This is since the
observations can be regarded as being generated from a dynamic mixture
model \cite{VanGael09} as in (\ref{marginalHMM}), the Dirichlet priors
on the rows have a shared parameter.
\begin{eqnarray}
p(w_i|h_{i-1}=k) &=& \sum_{h_i=1}^K p(h_i | h_{i-1}=k) p(w_i|h_i)\nonumber\\
&=&\sum_{h_i=1}^K \pi_{k,h_i} p(w_i | \phi_{h_i})
\label{marginalHMM}
\end{eqnarray}

In the right side in Figure \ref{ngramHMMFigure}, we place Pitman-Yor prior
$\mbox{PY}$, which has advantage in its power-law behavior as our target is
NLP, as in (\ref{PY}):
\begin{eqnarray}
w_{i} | w_{1:i-1} &\sim& \mbox{PY}(d_i,\theta_i,G_i)
\label{PY}
\end{eqnarray}
where $\alpha$ is a concentration parameter, $\theta$ is a strength parameter,
and $G_i$ is a base measure. This construction is borrowed from hierarchical
Pitman-Yor language model \cite{Teh06}. 

\paragraph{Inference}

We compute the expected value of the posterior distribution of the hidden
variables with a beam search \cite{VanGael09}. This blocked Gibbs sampler
alternate samples the parameters (transition matrix, output parameters), the
state sequence, hyper-parameters, and the parameters related to language model
smoothing. As is mentioned in \cite{VanGael09}, this sampler has
characteristic in that it adaptively truncates the state space and run dynamic
programming as in (\ref{dynamic}):
\begin{eqnarray}
p(h_t|w_{1:t},u_{1:t})&=&p(w_t|h_t) \sum_{h_{t-1}:u_t < \pi^{(h_{t-1},h_t)}} p(h_{t-1}|w_{1:t-1},u_{1:t-1})
\label{dynamic}
\end{eqnarray}
where $u_t$ is only valid if this is smaller than the transition probabilities
of the hidden word sequence $h_1,\ldots,h_K$.  Note that we use an auxiliary
variable $u_i$ which samples for each word in the sequence from the
distribution $u_i \sim \mbox{Uniform}(0, \pi^{(h_{i-1},h_i)})$.  The
implementation of the beam sampler consists of preprocessing the transition
matrix $\pi$ and sorting its elements in descending order. 

\paragraph{Initialization}

First, we obtain the parameters for hierarchical Pitman-Yor process-based
language model \cite{Teh06,Goldwater06}, which can be obtained using a block
Gibbs sampling \cite{Mochihashi09}.

Second, in order to obtain a better initialization value $h$ for the
above inference, we perform the following EM algorithm instead of
giving the distribution of $h$ randomly. This EM algorithm
incorporates the above mentioned truncation \cite{VanGael09}.  In the
E-step, we compute the expected value of the posterior distribution of
the hidden variables.  For every position $h_i$, we send a forward
message $\alpha(h_{i-n+1:i-1})$ in a single path from the start to the
end of the chain (which is the standard forward recursion in HMM;
Hence we use $\alpha$). Here we normalize the sum of $\alpha$
considering the truncated variables $u_{i-n+1:i-1}$.
\begin{eqnarray}
\alpha(h_{i-n+2:i}) &=& \frac{\sum \alpha(h_{i-n+1:i-1})}{\sum
  \alpha(u_{i-n+1:i-1})}P(w_i|h_i) \sum \alpha(u_{i-n+1:i-1})P(h_i|h_{i-n+1:i-1})
\end{eqnarray}
Then, for every position $h_j$, we send a message $\beta(h_{i-n+2:i},h_j)$ in
multiple paths from the start to the end of the chain as in
(\ref{secondForward}),
\begin{eqnarray}
\beta(h_{i-n+2:i},h_j) &=& 
\frac{\sum \alpha(h_{i-n+1:i-1})}{\sum \alpha(u_{i-n+1:i-1})} P(w_i|h_i)
\sum \beta(h_{i-n+1:i-1},h_j) P(h_i|h_{i-n+1:i-1})
\label{secondForward}
\end{eqnarray}
This step aims at obtaining the expected value of the posterior
distribution (Similar construction to use expectation can be seen in
factored HMM \cite{Ghahramani97}). In the M-step, using this expected
value of the posterior distribution obtained in the E-step to evaluate
the expectation of the logarithm of the complete-data likelihood.

\section{Joint Space Model}

\begin{figure}[h]
\begin{center}
\includegraphics[width=4cm]{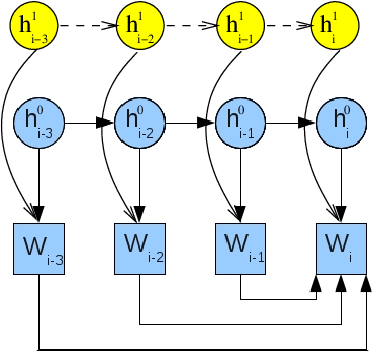}
\end{center}
\caption{Figure shows the joint space 4-gram HMM language model.}
\label{joint}
\end{figure}

In this paper, we mechanically introduce a joint space model. Other than the
ngram-HMM language model obtained in the previous section, we will often
encounter the situation where we have another hidden variables $h^1$ which is
irrelevant to $h^0$ which is depicted in Figure \ref{joint}.  Suppose that we
have the ngram-HMM language model yielded the hidden variables suggesting
semantic and syntactical role of words. Adding to this, we may have another
hidden variables suggesting, say, a genre ID. This genre ID can be considered
as the second context which is often not closely related to the first
context. This also has an advantage in this mechanical construction that the
resulted language model often has the perplexity smaller than the original
ngram-HMM language model.  Note that we do not intend to learn this model
jointly using the universal criteria, but we just concatenate the labels by
different tasks on the same sequence. By this formulation, we intend to
facilitate the use of this language model.

It is noted that those two contexts may not be derived in a single
learning algorithm. For example, language model with the sentence
context may be derived in the same way with that with the word
context. In the above example, a hidden semantics over sentence is not
a sequential object. Hence, this can be only considering all the
sentence are independent. Then, we can obtain this using, say, LDA.

\section{Intrinsic Evaluation}

We compared the perplexity of ngram-HMM LM (1 feature), ngram-HMM LM
(2 features, the same as in this paper and genre ID is 4 class),
modified Kneser-Ney smoothing (irstlm) \cite{Federico10}, and
hierarchical Pitman Yor LM \cite{Teh06}. We used news2011 English
testset. We trained LM using Europarl.

\begin{table*}[h] 
\begin{center}
\begin{tabular}{lllll} \hline
&ngram-HMM (1 feat) & ngram-HMM (2 feat) & modified Kneser-Ney & hierarchical PY \\ \hline
Europarl 1500k &114.014 & 113.450 & 118.890 & 118.884 \\ \hline
\end{tabular}
\end{center}
\caption{Table shows the perplexity of each language model.}
\label{perplexity}
\end{table*}

\section{Extrinsic Evaluation: Task of System Combination}

We applied ngram-HMM language model to the task of system
combination. For given multiple Machine Translation (MT) outputs, this
task essentially combines the best fragments among given MT outputs to
recreate a new MT output. The standard procedure consists of three
steps: Minimum Bayes Risk decoding, monolingual word alignment, and
monotonic consensus decoding. Although these procedures themselves
will need explanations in order to understand the following, we keep
the main text in minimum, moving some explanations (but not
sufficient) in appendices. Note that although this experiment was done
using the ngram-HMM language model, any \textsc{NPLM} may be
sufficient for this purpose. In this sense, we use the term
\textsc{NPLM} instead of ngram-HMM language model.


\paragraph{Features in Joint Space}
\label{LDA}

The first feature of \textsc{NPLM} is the semantically and
syntactically similar words of roles, which can be derived from the
original \textsc{NPLM}.  We introduce the second feature in this
paragraph, which is a genre ID.

The motivation to use this feature comes from the study of domain
adaptation for SMT where it becomes popular to consider the effect of
genre in testset. This paper uses Latent Dirichlet Allocation (LDA)
\cite{Blei03,Steyvers07,blei2011introduction,Sontag11,Murphy12} to
obtain the genre ID via (unsupervised) document classification since
our interest here is on the genre of sentences in testset.  And then,
we place these labels on a joint space.

LDA represents topics as multinomial distributions over the $W$ unique
word-types in the corpus and represents documents as a mixture of
topics. Let $C$ be the number of unique labels in the corpus. Each
label $c$ is represented by a $W$-dimensional multinomial distribution
$\phi_c$ over the vocabulary. For document $d$, we observe both the
words in the document $w^{(d)}$ as well as the document labels
$c^{(d)}$. Given the distribution over topics $\theta_d$, the
generation of words in the document is captured by the following
generative model.The parameters $\alpha$ and $\beta$ relate to the
corpus level, the variables $\theta_d$ belong to the document level,
and finally the variables $z_{dn}$ and $w_{dn}$ correspond to the word
level, which are sampled once for each word in each document. 

Using topic modeling in the second step, we propose the overall
algorithm to obtain genre IDs for testset as in (\ref{AlgorithmTopic}).

\begin{enumerate}
\label{AlgorithmTopic}
\item Fix the number of clusters $C$, we explore values from small to
  big where the optimal value will be searched on tuning set.
\item Do unsupervised document classification (or LDA) on the source side
of the tuning and test sets.
\begin{enumerate}
\item For each label $c \in \{1,\ldots C\}$, sample a distribution over
word-types $\phi_c \sim \mbox{\bf Dirichlet}(\cdot|\beta)$
\item For each document $d \in \{1,\ldots,D \}$
\begin{enumerate}
\item Sample a distribution over its observed labels $\theta_d \sim \mbox{\bf Dirichlet}(\cdot|\alpha)$
\item For each word $i \in \{1,\ldots,N_d^W\}$ 
\begin{enumerate}
\item Sample a label $z_i^{(d)} \sim \mbox{\bf Multinomial}(\theta_d)$
\item Sample a word $w_i^{(d)} \sim \mbox{\bf Multinomial}(\phi_c)$ from the label $c=z_i^{(d)}$
\end{enumerate}
\end{enumerate}
\end{enumerate}
\item Separate each class of tuning and test sets (keep the original index and new index in the allocated separated dataset).
\item (Run system combination on each class.)
\item (Reconstruct the system combined results of each class preserving the original index.)
\end{enumerate}

\paragraph{Modified Process in System Combination}
\label{syscom}

Given a joint space of \textsc{NPLM}, we need to specify in which
process of the task of system combination among three processes use
this \textsc{NPLM}.  We only discuss here the standard system
combination using confusion-network.  This strategy takes the
following three steps (Very brief explanation of these three is
available in Appendix):
\begin{itemize}
\item Minimum Bayes Risk decoding \cite{Kumar02} (with Minimum Error Rate
  Training (MERT) process \cite{Och03})
\begin{eqnarray}
\lefteqn{\hat{E}_{best}^{MBR} = \mbox{arg} \mbox{min}_{E' \in \mathcal{E}} R(E') = \mbox{arg} \mbox{min}_{E' \in \mathcal{E}} \sum_{E' \in \mathcal{E}_E} L(E,E')P(E|F)} \nonumber\\
\label{MBRDecoder}
&=& \mbox{arg} \mbox{min}_{E' \in \mathcal{E}} \sum_{E' \in \mathcal{E}_E} (1-BLEU_E(E'))P(E|F) \nonumber 
\end{eqnarray}
\item Monolingual word alignment
\item (Monotone) consensus decoding (with MERT process)
\begin{eqnarray}
E_{best} &=& \arg \max_e \prod_{i=1}^I \phi(i | \bar{e}_i)  p_{LM}(e)
\nonumber
\end{eqnarray}
\end{itemize}

Similar to the task of n-best reranking in MERT process
\cite{Schwenk12}, we consider the reranking of nbest lists in the
third step of above, i.e.  (monotone) consensus decoding (with MERT
process). We do not discuss the other two processes in this paper.

On one hand, we intend to use the first feature of \textsc{NPLM},
i.e. the semantically and syntactically similar role of words, for
paraphrases.  The n-best reranking in MERT process \cite{Schwenk12}
alternate the probability suggested by word sense disambiguation task
using the feature of \textsc{NPLM}, while we intend to add a sentence
which replaces the words using \textsc{NPLM}. On the other hand, we
intend to use the second feature of \textsc{NPLM}, i.e. the genre ID,
to split a single system combination system into multiple system
combination systems based on the genre ID clusters.  In this
perspective, the role of these two feature can be seen as
independent. We conducted four kinds of settings below.

\paragraph{(A) ---First Feature: N-Best Reranking in Monotonic Consensus Decoding without Noise -- NPLM plain}

In the first setting for the experiments, we used the first feature without
considering noise. The original aim of \textsc{NPLM} is to capture the
semantically and syntactically similar words in a way that a latent word
depends on the context. We will be able to get variety of words if we
condition on the fixed context, which would form paraphrases in theory.

We introduce our algorithm via a word sense disambiguation (WSD) task
which selects the right disambiguated sense for the word in question.
This task is necessary due to the fact that a text is natively
ambiguous accommodating with several different meanings.  The task of
WSD \cite{Deschacht12} can be written as in (\ref{wsd}):
\begin{eqnarray}
P(\mbox{synset}_i|\mbox{features}_i, \theta) &=& \frac{1}{Z(\mbox{features})}
\prod_m g(\mbox{synset}_i,k)^{f(\mbox{feature}_i^k)}
\label{wsd}
\end{eqnarray}
where $k$ ranges over all possible features, $f(\mbox{feature}_i^k)$
is an indicator function whose value is 1 if the feature exists, and 0
otherwise, $g(\mbox{synset}_i,k)$ is a parameter for a given synset
and feature, $\theta$ is a collection of all these parameters in
$g(\mbox{synset}_i,k)$, and $Z$ is a normalization constant. Note that
we use the term ``synset'' as an analogy of the WordNet
\cite{Miller95}: this is equivalent to ``sense'' or ``meaning''. Note
also that \textsc{NPLM} will be included as one of the features in
this equation.  If features include sufficient statistics, a task of
WSD will succeed. Otherwise, it will fail. We do reranking of the
outcome of this WSD task.

On the one hand, the paraphrases obtained in this way have attractive
aspects that can be called ``a creative word'' \cite{Veale12}. This is
since the traditional resource that can be used when building a
translation model by SMT are constrained on parallel corpus. However,
\textsc{NPLM} can be trained on huge monolingual corpus. On the other
hand, unfortunately in practice, the notorious training time of
\textsc{NPLM} only allows us to use fairly small monolingual corpus
although many papers made an effort to reduce it \cite{MniTeh2012a}.
Due to this, we cannot ignore the fact that \textsc{NPLM} trained not
on a huge corpus may be affected by noise. Conversely, we have no
guarantee that such noise will be reduced if we train \textsc{NPLM} on
a huge corpus. It is quite likely that \textsc{NPLM} has a lot of
noise for small corpora. Hence, this paper also needs to provide the
way to
overcome difficulties of noisy data. In order to avoid this difficulty, we limit the paraphrase only when it includes itself in high
probability.

\paragraph{(B)--- First Feature: N-Best Reranking in Monotonic Consensus Decoding with Noise -- NPLM dep}

\begin{figure}[h]
\begin{center}
\includegraphics[scale=0.5]{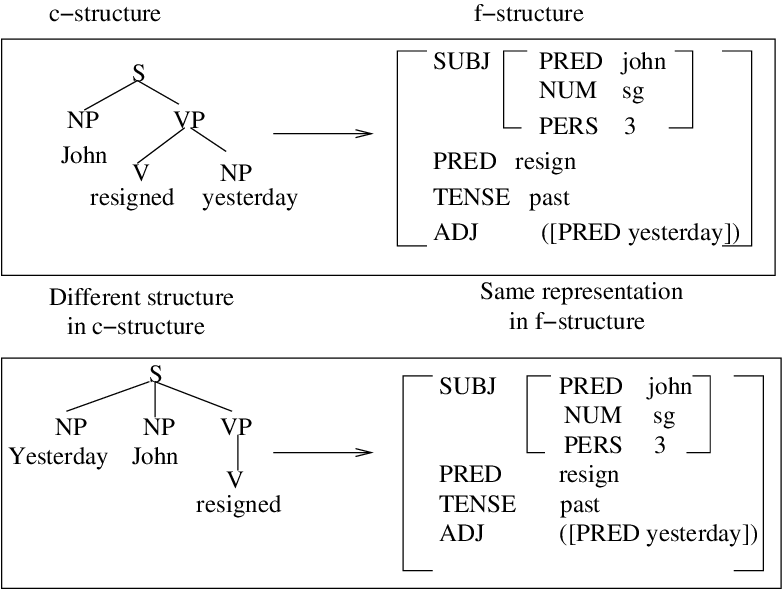}
\end{center}
\caption{By the modified dependency score \cite{Owczarzak07}, the
  score of these two sentences, ``John resigned yesterday'' and
  ``Yesterday John resigned'', are the same.  Figure shows c-structure
  and f-structure of two sentences using Lexical Functional Grammar
  (LFG) \cite{Bresnan01}.}
\label{John}
\end{figure}

In the second setting for our experiment, we used the first feature
considering noise. Although we modified a suggested paraphrase
without any intervention in the above algorithm, it is also possible
to examine whether such suggestion should be adopted or not.  If we
add paraphrases and the resulted sentence has a higher score in terms
of the modified dependency score \cite{Owczarzak07} (See Figure
\ref{John}), this means that the addition of paraphrases is a good
choice. If the resulted score decreases, we do not need to add
them. One difficulty in this approach is that we do not have a
reference which allows us to score it in the usual manner. For this
reason, we adopt the {\it naive way} to deploy the above and we deploy
this with {\it pseudo references}. (This formulation is equivalent
that we decode these inputs by MBR decoding.) First, if we add
paraphrases and the resulted sentence does not have a very bad score,
we add these paraphrases since these paraphrase are not very bad ({\it
  naive} way). Second, we do scoring between the sentence in question
with {\it all the other candidates} ({\it pseudo references}) and
calculate an average of them. Thus, our second algorithm is to select
a paraphrase which may not achieve a very bad score in terms of the
modified dependency score using \textsc{NPLM}.

\paragraph{(C) --- Second Feature: Genre ID --- DA (Domain Adaptation)}

In the third setting of our experiment, we used only the second feature.  As
is mentioned in the explanation about this feature, we intend to splits a
single module of system combination into multiple modules of system
combination according to the genre ID. Hence, we will use the module of system
combination tuned for the specific genre ID, \footnote{E.g., we translate
  newswire with system combination module tuned with newswire tuning set,
  while we translate medical text with system combination module tuned with
  medical text tuning set.}.

\paragraph{(D) --- First and Second Feature --- COMBINED}

In the fourth setting we used both features.  In this setting, (1) we used
modules of system combination which are tuned for the specific genre ID, and
(2) we prepared NPLM whose context can be switched based on the specific genre
of the sentence in test set. The latter was straightforward since these two
features are stored in joint space in our case.

\paragraph{Experimental Results}

ML4HMT-2012 provides four translation outputs (\textit{s1} to
\textit{s4}) which are MT outputs by two RBMT systems,
\textsc{apertium} and \textsc{Lucy}, \textsc{PB-SMT (Moses)} and
\textsc{HPB-SMT (Moses)}, respectively.  The tuning data consists of
20,000 sentence pairs, while the test data consists of 3,003 sentence
pairs.

Our experimental setting is as follows. We use our system combination
module \cite{Du10,Du10b,Okita2012}, which has its own language
modeling tool, MERT process, and MBR decoding. We use the
\textsc{BLEU} metric as loss function in MBR decoding. We use
\textsc{TERp}\footnote{{\it http://www.cs.umd.edu/~snover/terp}} as
alignment metrics in monolingual word alignment.  We trained \textsc{NPLM}
using 500,000 sentence pairs from English side of EN-ES corpus of
\textsc{Europarl}\footnote{{\it http://www.statmt.org/europarl}}.

The results show that the first setting of \textsc{NPLM}-based paraphrased
augmentation, that is \textsc{NPLM} plain, achieved 25.61 \textsc{BLEU}
points, which lost 0.39 \textsc{BLEU} points absolute over the standard system
combination.  The second setting, \textsc{NPLM} dep, achieved slightly better
results of 25.81 \textsc{BLEU} points, which lost 0.19 \textsc{BLEU} points
absolute over the standard system combination.  Note that the baseline
achieved 26.00 \textsc{BLEU} points, the best single system in terms of
\textsc{BLEU} was s4 which achieved 25.31 \textsc{BLEU} points, and the best
single system in terms of \textsc{METEOR} was s2 which achieved 0.5853.
The third setting achieved 26.33 \textsc{BLEU} points, which was the best
among our four settings. The fourth setting achieved 25.95, which is again
lost 0.05 \textsc{BLEU} points over the standard system combination.

Other than our four settings where these settings differ which features to
use, we run several different settings of system combination in order to
understand the performance of four settings. Standard system combination using
BLEU loss function (line 5 in Table 2), standard system combination using TER
loss function (line 6), system combination whose backbone is unanamously taken
from the RBMT outputs (MT input s2 in this case; line 11), and system
combination whose backbone is selected by the modified dependency score (which
has three variations in the figure; modDep precision, recall and Fscore; line
12, 13 and 14). One interesting characteristics is that the s2 backbone (line
11) achieved the best score among all of these variations. Then, the score of
the modified dependency measure-selected backbone follows. From these runs, we
cannot say that the runs related to \textsc{NPLM}, i.e. (A), (B) and (D), were
not particularly successful. The possible reason for this was that our
interface with \textsc{NPLM} was only limited to paraphrases, which was not
very successfuly chosen by reranking.

\begin{table*}[h] 
\begin{center}
\begin{tabular}{llllll} \hline
 &\textsc{NIST} & \textsc{BLEU} & \textsc{METEOR} & \textsc{WER} &\textsc{PER} \\ \hline
MT input s1&     6.4996  &0.2248 &0.5458641 &64.2452 &49.9806\\
MT input s2&     6.9281  &0.2500 &\underline{0.5853446} &62.9194 &48.0065\\
MT input s3&     7.4022  &0.2446 &0.5544660 &58.0752 &44.0221\\
MT input s4&     7.2100  &\underline{0.2531} &0.5596933 &59.3930 &44.5230\\\hline \hline
standard system combination (BLEU) &7.6846&  0.2600& 0.5643944& 56.2368& 41.5399\\
standard system combination (TER)   &7.6231&  0.2638& 0.5652795& 56.3967& 41.6092\\ \hline
(A) \textsc{NPLM} plain &7.6041 &0.2561 &0.5593901 &56.4620 &41.8076  \\
(B) \textsc{NPLM} dep   &7.6213 &0.2581 &0.5601121 &56.1334 &41.7820 \\
(C) DA &7.7146&  0.2633& 0.5647685& 55.8612& 41.7264\\ 
(D) COMBINED & 7.6464& 0.2595& 0.5610121& 56.0101 & 41.7702\\
\hline
s2 backbone &7.6371&  \underline{0.2648}& 0.5606801& 56.0077& 42.0075\\
modDep precision &7.6670& 0.2636& 0.5659757& 56.4393& 41.4986\\
modDep recall &7.6695  &0.2642 &0.5664320 &56.5059 &41.5013 \\
modDep Fscore &7.6695  &0.2642 &0.5664320 &56.5059 &41.5013 \\ \hline\hline
\end{tabular}
\end{center}
\caption{This table shows single best performance, the performance of the
  standard system combination (BLEU and TER loss functions), the performance
  of four settings in this paper ((A),$\ldots$,(D)), the performance of s2
  backboned system combination, and the performance of the selection of
  sentences by modified dependency score (precision, recall, and F-score
  each). }
\label{modDepBackbone}
\end{table*}

\section*{Conclusion and Perspectives} 

This paper proposes a non-parametric Bayesian way to interpret \textsc{NPLM},
which we call ngram-HMM language model. Then, we add a small extension to this
by concatenating other context in the same model, which we call a joint space
ngram-HMM language model. The main issues investigated in this paper was an
application of \textsc{NPLM} in bilingual NLP, specifically Statistical
Machine Translation (SMT). We focused on the perspectives that \textsc{NPLM}
has potential to open the possibility to complement potentially `huge'
monolingual resources into the `resource-constraint' bilingual resources.  We
compared our proposed algorithms and others. One discovery was that when we
use a fairly small \textsc{NPLM}, noise reduction may be one way to improve
the quality. In our case, the noise reduced version obtained 0.2 BLEU points
better.

Further work would be to apply this \textsc{NPLM} in various other tasks in
SMT: word alignment, hierarchical phrase-based decoding, and semantic
incorporated MT systems in order to discover the merit of `depth'
of architecture in Machine Learning. 

{\tiny
\bibliographystyle{acm}
\bibliography{nplm}}

\begin{thebibliography}{10}

\bibitem{Bangalore02}
{\sc Bangalore, S., Bordel, G., and Riccardi, G.}
\newblock Computing consensus translation from multiple machine translation
  systems.
\newblock {\em In Proceedings of the IEEE Automatic Speech Recognition and
  Understanding Workshop (ASRU)\/} (2001), 350--354.

\bibitem{Beal03}
{\sc Beal, M.~J.}
\newblock Variational algorithms for approximate bayesian inference.
\newblock {\em PhD Thesis at Gatsby Computational Neuroscience Unit, University
  College London\/} (2003).

\bibitem{Bengio00}
{\sc Bengio, Y., Ducharme, R., and Vincent, P.}
\newblock A neural probabilistic language model.
\newblock {\em In Proceedings of Neural Information Systems\/} (2000).

\bibitem{Bengio05}
{\sc Bengio, Y., Schwenk, H., Sen\'ecal, J.-S., Morin, F., and Gauvain, J.-L.}
\newblock Neural probabilistic language models.
\newblock {\em Innovations in Machine Learning: Theory and Applications Edited
  by D.~Holmes and L.~C.~Jain\/} (2005).

\bibitem{Blei03}
{\sc Blei, D., Ng, A.~Y., and Jordan, M.~I.}
\newblock Latent dirichlet allocation.
\newblock {\em Journal of Machine Learning Research 3\/} (2003), 993–1022.

\bibitem{blei2011introduction}
{\sc Blei, D.~M.}
\newblock Introduction to probabilistic topic models.
\newblock {\em Communications of the ACM\/} (2011).

\bibitem{Bordes11}
{\sc Bordes, A., Glorot, X., Weston, J., and Bengio, Y.}
\newblock Towards open-text semantic parsing via multi-task learning of
  structured embeddings.
\newblock {\em CoRR abs/1107.3663\/} (2011).

\bibitem{Bresnan01}
{\sc Bresnan, J.}
\newblock Lexical functional syntax.
\newblock {\em Blackwell\/} (2001).

\bibitem{Chen98}
{\sc Chen, S., and Goodman, J.}
\newblock An empirical study of smoothing techniques for language modeling.
\newblock {\em Technical report TR-10-98 Harvard University\/} (1998).

\bibitem{Collobert11a}
{\sc Collobert, R.}
\newblock Deep learning for efficient discriminative parsing.
\newblock {\em In Proceedings of the 14th International Conference on
  Artificial Intelligence and Statistics (AISTATS)\/} (2011).

\bibitem{Collobert08}
{\sc Collobert, R., and Weston, J.}
\newblock A unified architecture for natural language processing: Deep neural
  networks with multitask learning.
\newblock {\em In International Conference on Machine Learning (ICML 2008)\/}
  (2008).

\bibitem{Collobert11b}
{\sc Collobert, R., Weston, J., Bottou, L., Karlen, M., Kavukcuoglu, K., and
  Kuksa, P.}
\newblock Natural language processing (almost) from scratch.
\newblock {\em Journal of Machine Learning Research 12\/} (2011), 2493--2537.

\bibitem{DeNero09}
{\sc DeNero, J., Chiang, D., and Knight, K.}
\newblock Fast consensus decoding over translation forests.
\newblock {\em In proceedings of the Joint Conference of the 47th Annual
  Meeting of the ACL and the 4th International Joint Conference on Natural
  Language Processing of the AFNLP\/} (2009), 567--575.

\bibitem{Deschacht12}
{\sc Deschacht, K., Belder, J.~D., and Moens, M.-F.}
\newblock The latent words language model.
\newblock {\em Computer Speech and Language 26\/} (2012), 384--409.

\bibitem{Du09b}
{\sc Du, J., He, Y., Penkale, S., and Way, A.}
\newblock {MaTrEx}: the {DCU} {MT} {System} for {WMT} 2009.
\newblock {\em In Proceedings of the Third EACL Workshop on Statistical Machine
  Translation\/} (2009), 95--99.

\bibitem{Du10}
{\sc Du, J., and Way, A.}
\newblock An incremental three-pass system combination framework by combining
  multiple hypothesis alignment methods.
\newblock {\em International Journal of Asian Language Processing 20}, 1
  (2010), 1--15.

\bibitem{Du10b}
{\sc Du, J., and Way, A.}
\newblock Using terp to augment the system combination for smt.
\newblock {\em In Proceedings of the Ninth Conference of the Association for
  Machine Translation (AMTA2010)\/} (2010).

\bibitem{Federico10}
{\sc Federico, M., Bertoldi, N., and Cettolo, M.}
\newblock Irstlm: an open source toolkit for handling large scale language
  models.
\newblock {\em Proceedings of Interspeech\/} (2008).

\bibitem{VanGael09}
{\sc Gael, J.~V., Vlachos, A., and Ghahramani, Z.}
\newblock The infinite hmm for unsupervised pos tagging.
\newblock {\em The 2009 Conference on Empirical Methods on Natural Language
  Processing (EMNLP 2009)\/} (2009).

\bibitem{Gasthaus10}
{\sc Gasthaus, J., Wood, F., and Teh, Y.~W.}
\newblock Lossless compression based on the sequence memoizer.
\newblock {\em DCC 2010\/} (2010).

\bibitem{Gharamani01}
{\sc Ghahramani, Z.}
\newblock An introduction to hidden markov models and bayesian networks.
\newblock {\em International Journal of Pattern Recognition and Artificial
  Intelligence 15}, 1 (2001), 9--42.

\bibitem{Ghahramani97}
{\sc Ghahramani, Z., Jordan, M.~I., and Smyth, P.}
\newblock Factorial hidden markov models.
\newblock {\em Machine Learning\/} (1997).

\bibitem{Goldwater06}
{\sc Goldwater, S., Griffiths, T.~L., and Johnson, M.}
\newblock Contextual dependencies in unsupervised word segmentation.
\newblock {\em In Proceedings of Conference on Computational Linguistics /
  Association for Computational Linguistics (COLING-ACL06)\/} (2006), 673--680.

\bibitem{Good53}
{\sc Good, I.~J.}
\newblock The population frequencies of species and the estimation of
  population paramters.
\newblock {\em Biometrika 40}, (3-4) (1953), 237--264.

\bibitem{Hinton86}
{\sc Hinton, G.~E., McClelland, J.~L., and Rumelhart, D.}
\newblock Distributed representations.
\newblock {\em Parallel Distributed Processing: Explorations in the
  Microstructure of Cognition(Edited by D.E.~Rumelhart and J.L.~McClelland) MIT
  Press 1\/} (1986).

\bibitem{Kneser95}
{\sc Kneser, R., and Ney, H.}
\newblock Improved backing-off for n-gram language modeling.
\newblock {\em In Proceedings of the IEEE International Conference on
  Acoustics, Speech and Signal Processing\/} (1995), 181--184.

\bibitem{Koller09}
{\sc Koller, D., and Friedman, N.}
\newblock Probabilistic graphical models: Principles and techniques.
\newblock {\em MIT Press\/} (2009).

\bibitem{Kumar02}
{\sc Kumar, S., and Byrne, W.}
\newblock Minimum {Bayes-Risk} word alignment of bilingual texts.
\newblock {\em In Proceedings of the Empirical Methods in Natural Language
  Processing (EMNLP 2002)\/} (2002), 140--147.

\bibitem{Matusov06}
{\sc Matusov, E., Ueffing, N., and Ney, H.}
\newblock Computing consensus translation from multiple machine translation
  systems using enhanced hypotheses alignment.
\newblock {\em In Proceedings of the 11st Conference of the European Chapter of
  the Association for Computational Linguistics (EACL)\/} (2006), 33--40.

\bibitem{Miller95}
{\sc Miller, G.~A.}
\newblock Wordnet: A lexical database for english.
\newblock {\em Communications of the ACM 38}, 11 (1995), 39--41.

\bibitem{MniTeh2012a}
{\sc Mnih, A., and Teh, Y.~W.}
\newblock A fast and simple algorithm for training neural probabilistic
  language models.
\newblock In {\em Proceedings of the International Conference on Machine
  Learning\/} (2012).

\bibitem{Mochihashi09}
{\sc Mochihashi, D., Yamada, T., and Ueda, N.}
\newblock Bayesian unsupervised word segmentation with nested pitman-yor
  language modeling.
\newblock {\em In Proceedings of Joint Conference of the 47th Annual Meeting of
  the Association for Computational Linguistics and the 4th International Joint
  Conference on Natural Language Processing of the Asian Federation of Natural
  Language Processing (ACL-IJCNLP 2009)\/} (2009), 100--108.

\bibitem{Murphy12}
{\sc Murphy, K.~P.}
\newblock Machine learning: A probabilistic perspective.
\newblock {\em The MIT Press\/} (2012).

\bibitem{Och03}
{\sc Och, F., and Ney, H.}
\newblock A systematic comparison of various statistical alignment models.
\newblock {\em Computational Linguistics 29}, 1 (2003), 19--51.

\bibitem{Okita2012}
{\sc Okita, T., and van Genabith, J.}
\newblock Minimum bayes risk decoding with enlarged hypothesis space in system
  combination.
\newblock {\em In Proceedings of the 13th International Conference on
  Intelligent Text Processing and Computational Linguistics (CICLING 2012).
  LNCS 7182 Part II. A. Gelbukh (Ed.)\/} (2012), 40--51.

\bibitem{Okita10e}
{\sc Okita, T., and Way, A.}
\newblock Hierarchical pitman-yor language model in machine translation.
\newblock {\em In Proceedings of the International Conference on Asian Language
  Processing (IALP 2010)\/} (2010).

\bibitem{Okita11}
{\sc Okita, T., and Way, A.}
\newblock {P}itman-{Y}or process-based language model for {M}achine
  {T}ranslation.
\newblock {\em International Journal on Asian Language Processing 21}, 2
  (2010), 57--70.

\bibitem{Okita11c}
{\sc Okita, T., and Way, A.}
\newblock Given bilingual terminology in statistical machine translation:
  Mwe-sensitve word alignment and hierarchical pitman-yor process-based
  translation model smoothing.
\newblock {\em In Proceedings of the 24th International Florida Artificial
  Intelligence Research Society Conference (FLAIRS-24)\/} (2011), 269--274.

\bibitem{Owczarzak07}
{\sc Owczarzak, K., van Genabith, J., and Way, A.}
\newblock Evaluating machine translation with {LFG} dependencies.
\newblock {\em Machine Translation 21}, 2 (2007), 95--119.

\bibitem{Rabiner89}
{\sc Rabiner, L.~R.}
\newblock A tutorial on hidden markov models and selected applications in
  speech recognition.
\newblock {\em Proceedings of the IEEE 77}, 2 (1989), 257--286.

\bibitem{Schwenk07}
{\sc Schwenk, H.}
\newblock Continuous space language models.
\newblock {\em Computer Speech and Language 21\/} (2007), 492--518.

\bibitem{Schwenk10}
{\sc Schwenk, H.}
\newblock Continuous space language models for statistical machine translation.
\newblock {\em The Prague Bulletin of Mathematical Linguistics 83\/} (2010),
  137--146.

\bibitem{Schwenk12}
{\sc Schwenk, H., Rousseau, A., and Attik, M.}
\newblock Large, pruned or continuous space language models on a gpu for
  statistical machine translation.
\newblock {\em In Proceeding of the NAACL workshop on the Future of Language
  Modeling\/} (2012).

\bibitem{Sontag10}
{\sc Sontag, D.}
\newblock Approximate inference in graphical models using {LP} relaxations.
\newblock {\em Massachusetts Institute of Technology (Ph.D. thesis)\/} (2010).

\bibitem{Sontag11}
{\sc Sontag, D., and Roy, D.~M.}
\newblock The complexity of inference in latent dirichlet allocation.
\newblock {\em In Advances in Neural Information Processing Systems 24
  (NIPS)\/} (2011).

\bibitem{Steyvers07}
{\sc Steyvers, M., and Griffiths, T.}
\newblock Probabilistic topic models.
\newblock {\em Handbook of Latent Semantic Analysis.~Psychology Press\/}
  (2007).

\bibitem{Stolcke02}
{\sc Stolcke, A.}
\newblock {SRILM} -- {A}n extensible language modeling toolkit.
\newblock {\em In Proceedings of the International Conference on Spoken
  Language Processing\/} (2002), 901--904.

\bibitem{Teh06}
{\sc Teh, Y.~W.}
\newblock A hierarchical bayesian language model based on pitman-yor processes.
\newblock {\em In Proceedings of the 44th Annual Meeting of the Association for
  Computational Linguistics (ACL-06), Prague, Czech Republic\/} (2006),
  985--992.

\bibitem{Tromble08}
{\sc Tromble, R., Kumar, S., Och, F., and Macherey, W.}
\newblock Lattice minimum bayes-risk decoding for statistical machine
  translation.
\newblock {\em Proceedings of the 2008 Conference on Empirical Methods in
  Natural Language Processing\/} (2008), 620--629.

\bibitem{Veale12}
{\sc Veale, T.}
\newblock Exploding the creativity myth: The computational foundations of
  linguistic creativity.
\newblock {\em London: Bloomsbury Academic\/} (2012).

\bibitem{Wood09}
{\sc Wood, F., Archambeau, C., Gasthaus, J., James, L., and Teh, Y.~W.}
\newblock A stochastic memoizer for sequence data.
\newblock {\em In Proceedings of the 26th International Conference on Machine
  Learning\/} (2009), 1129--1136.

\end{thebibliography}

\end{document}